\documentclass{article}

\usepackage[preprint]{corl_2023} 

\usepackage{float}
\usepackage{amsmath}
\usepackage{amssymb}
\usepackage{mathtools}
\usepackage{dsfont}
\usepackage{amsthm}
\usepackage{listings}
\usepackage{algorithm}
\usepackage{algpseudocode}

\usepackage{enumitem}
\def\digits#1{%
  \number#1}

\newcommand{\learnedR}{\ensuremath{R_{\theta}}}
\newcommand{\state}{\ensuremath{\mathcal{S}}}
\newcommand{\action}{\ensuremath{\mathcal{A}}}
\newcommand{\reward}{\ensuremath{\mathcal{R}}}
\newcommand{\transition}{\ensuremath{\mathcal{T}}}
\newcommand{\horizon}{\ensuremath{T}}
\newcommand{\traj}[1]{\ensuremath{\tau}_{#1}}
\newcommand{\dataset}{\ensuremath{\mathcal{D}}}
\newcommand{\loss}{\ensuremath{\mathcal{L}}}
\newcommand{\replay}{\ensuremath{\mathcal{D}_{RB}}}
\newcommand{\rank}[1]{\ensuremath{Rank_{#1}}}
\newcommand{\hypo}[1]{\textit{\textbf{H#1}}}

\title{Learning Reward for Physical Skills using Large Language Model}

\author{
  Yuwei Zeng
  \quad
  Yiqing Xu \\
  School of Computing  \\
  National University of Singapore\\
}

\begin{document}
\maketitle


\begin{abstract}
Learning reward functions for physical skills are challenging due to the vast spectrum of skills, the high-dimensionality of  state and action space, and nuanced sensory feedback. The complexity of these tasks makes acquiring expert demonstration data both costly and time-consuming. Large Language Models (LLMs) contain valuable task-related knowledge that can aid in learning these reward functions. However, the direct application of LLMs for proposing reward functions has its limitations such as numerical instability and inability to incorporate the environment feedback.
We aim to extract task knowledge from LLMs using environment feedback to create efficient reward functions for physical skills. Our approach consists of two components. We first use the LLM to propose features and parameterization of the reward function. Next, we update the parameters of this proposed reward function through an iterative self-alignment process.  In particular, this process minimizes the ranking inconsistency between the LLM and our learned reward functions based on the new observations. We validated our method by testing it on three simulated physical skill learning tasks, demonstrating effective support for our design choices.
\end{abstract}



\section{Introduction}
\label{sec:intro}

Robotics has advanced significantly in optimizing complex physical skills, from dynamic walking over uneven terrain \cite{uneven_terrain, manchester2011stable, kumar2021rma} to high-precision peg-in-hole assembly \cite{xu2019compare, li2022robotic, 9811265}. The effectiveness of these methods largely depends on a well-structured reward function. However, creating these reward functions manually is time-consuming and necessitates extensive expert knowledge - a significant barrier to the broad application of RL and motion planning methods in various physical skills \cite{IM_book, IRL_survey}. Inverse reinforcement learning (IRL) seeks to resolve this issue by automatically learning the reward function from expert demonstrations \cite{AndrewNg, Apprenticeship}. Yet, due to the vast variety and complexity of the state space for physical tasks and the precision required for execution, gathering expert demonstrations can be exceedingly costly and gaining sufficient coverage is difficult. Hence, learning reward functions for a repertoire of physical skills presents a challenge. 

Large Languaged Models (LLMs), trained using extensive human data, have proven to be effective in embedding useful task-related knowledge. For instance, we can query the LLM to translate a physical skill such as ``touch the block'' into a sequence of rules based on the observation signals. These rules include steps like ``move closer to the block'' and ``slow down when near the block''. This alternate method of acquiring task-specific information eliminates the need for time-consuming expert data collection. Several existing studies have successfully used LLM to directly propose or predict reward functions for the subsequent task planning LLM reward feedback \cite{kwon2023reward, yu2023language, adeniji2023language, xie2023text2reward}. Nevertheless, determining the reward function for physical skills remains a challenging area.

Using LLM to learn reward functions for physical skills presents a significant challenge due to the skills' sensitivity to the exact numerical values of the reward function. The limited capacity of LLM in the numerical reasoning makes directly proposing the reward functions less feasible. Some works use LLM to mimic human behavior and generate demonstration data to learn the reward function. This involves describing tasks in natural language and prompting LLM to create a series of actions and predicted states. However, due to the complexity of transition functions in physical skills, LLM often lacks the ability to accurately predict low-level controls and resulting states, rendering this method ineffective. Yet, LLM can yield insights on what are the important observation signals to watch out or the trends towards task completion. This begs the question: is there a more adaptable, controlled way to extract this knowledge from LLM for learning a reward function?

In this work, we propose a iterative self-alignment method to reward learning of physical skills using LLM. We first utilize the highly abstracted task understanding embedded in LLM to first craft the reward function parameterization, particularly the feature selection and template structure, eliminating the need for manual engineering. Next, our iterative self-alignment reward learning method operates on a double-loop structure to update the parameters of the proposed reward function. In particular, the inner loop induces the optimal policy from the current reward function, samples trajectories using this policy, and adds them to the replay buffer. The outer loop updates the parameters of the learned reward function by aligning the trajectory ranking based on the current reward function with the LLM proposed ranking. This process is similar to IRL's standard bi-level optimization structure, with one key difference in the outer loop: instead of minimizing differences between expert demonstrations and induced trajectories as in IRL \cite{f_D}, our method employs ranking as an alternative due to LLMs' inability to generate demonstrations. This way we extract task preferences from LLMs for comparison. 
Since all supervision signals, both direct (asking the LLM for ranking) or indirect (using the reward function learned from the LLM for calculation and ranking), come from the LLM, we describe this as the self-alignment reward update.

In summary, we introduce a self-alignment method for learning the reward function of physical skills from LLM. With no expert demonstration or feedback, our methods efficiently learn an accurate reward function for the physical skills that require fine-grained sensory input and precise controls. We evaluated our approach on three simulated tasks and demonstrate that our learned reward function can encourage optimal behaviors with significantly fewer training steps compared to the baselines.

\section{Related work}
\label{sec:related_work}

\paragraph{LLM for robotics reward learning}
Large Language Models (LLMs) have shown great potential as cost-effective tools for extracting task objectives as they embed substantial human knowledge. Various studies have utilized LLMs to predict reward functions \cite{kwon2023reward, yu2023language, adeniji2023language, xie2023text2reward}, although some methodologies hold limitations. For example, some use LLMs in a zero-shot approach
\cite{huang2023voxposer, adeniji2023language}, implying their reward function can't utilize environmental feedback. As a result, these models are limited to tasks requiring simple semantic reasoning due to their inability to understanding the dynamics of the environment. Other research focuses on reward learning for physical skills, a process sensitive to the reward function's numerical values. These studies typically require manual validation and reward function updating per task or rely on human feedback for supervising reward learning \cite{kwon2023reward, yu2023language, xie2023text2reward}, which can be laborious. Our research focuses on learning reward functions for physical skills. Instead of depending on human feedback, we suggest an iterative self-alignment that uses LLMs to offer alternative signals that supervise reward learning.

\paragraph{Inverse Reinforcement Learning from various forms of Human Data}
Inverse reinforcement learning (IRL) studies how to autonomously learn a reward function from expert data \cite{GPIRL, maxentirl, DeepMaxEnt, GCL, GAN-GCL, AIRL, EAIRL, GAIL, FAIRL, REIRL, fIRL, xu2022receding, BIRL, IRL_PG, opirl, maxmargin_IRL, f_D}. Traditional IRL methods utilise the optimal expert demonstrations as supervision signals, aiming to learn a reward function that can encode an objective consistent with the expert demonstrations. The bottleneck of IRL is the quantity and quality of the expert demonstrations. Some other works explore other forms of the expert supervision, for example, using sub-optimal practice data, failure trials \cite{SSRR, brown2019extrapolating}, or rankings of the trajectories. 

An alternative path is to learn from trajectory preference/ranking. These methods offer some benefits over standard IRL, which relies on almost perfect expert demonstrations for imitation. Essentially, this approach learns from ranked sub-optimal demonstrations, potentially surpassing the original demonstrator's performance \cite{hadfield2017inverse, lee2021pebble, lee2021b, mehta2022unified}. This makes it less demanding for experts to create only optimal behaviours, and it also encodes further contrasting information about preferable and undesirable behaviours. A major downside, however, remains the need to amass this information. Collecting such data can be labour-intensive, and potentially cost-prohibitive, when trying to learn reward functions across numerous physical skills. We investigate the potential to extract task-related data from a pre-trained large language model, effectively bypassing this data collection challenge.

\section{Main Method}
\label{sec:main}

\subsection{Problem Formulation}

Consider a finite-horizon Markov Decision Process (MDP) parameterized by $(\state, \action, \transition, \reward, \horizon)$ where
$\state$, $\action$ are the state and action spaces, $\transition: \state \times \action \to \state$ is the transition function, $\reward: \state \to \mathbb{R}$ is the reward
function, and $\horizon$ is the horizon. A policy $\pi$ is a mapping from states to probabilities over actions, $\pi(a|s)$. Given a
policy and an MDP, the expected return of the policy is given by $J(\pi;\reward) = \mathbb{E}[\sum_{t=0}^{T-1}\reward(s_t)|\pi]$. The expert policy should be one that optimizes this return, $\pi_E := \arg\max J(\pi; \reward)$ w.r.t. the ground-truth reward $\reward$. In our setting, we are given a partial MDP without i) the reward function $\reward$ nor ii) any forms of expert demonstrations. Instead, we have access to an LLM that can rank a sequence of $M$ trajectories $\traj{k}$ with decreasing preference for $k = 1, \ldots, m$ based on the last state $s^k_T$. We assume that the LLM has
an internal goal or intrinsic understanding of tasks, therefore its ranking is consistent with a human demonstrator optimizing the ground-truth reward function $\reward$.  We aim to find a parameterized reward function $\learnedR$ such that the ranking of these $M$ trajectories based on it is consistent with the ranking given by the LLM. 

\subsection{Method}

\begin{figure}[]
\centering
    \includegraphics[width=1\linewidth]{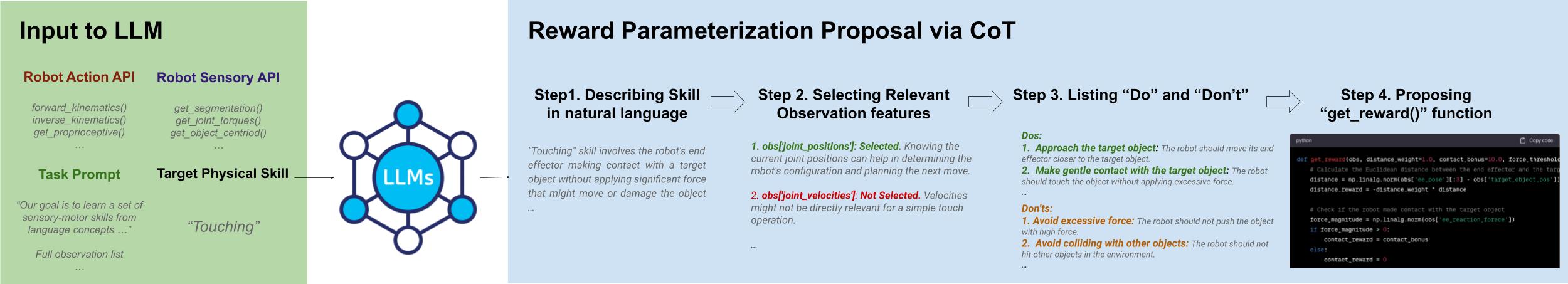} 
\caption{\small \textbf{Illustration for Reward Parameterization Proposal}. We designed a four-step procedure for proposing a reward parameterization using the Chain-of-Thought (CoT) method. We first input robot APIs and task specifications into the LLM to ground the desired physical skill in the context of this specific robotic system. We provide the step-wise sample outputs for the \lstinline{touching} task. Refer to the appendix for complete prompts.}
\label{fig:para_proposal}
\end{figure}

We propose a method to learn the reward functions for physical skills from Large Language Models (LLMs). We observe that LLM not only encodes useful information on completing the physical skills, it can also serve as a discriminator to evaluate the task performance given the observation signals. Based on these insights, our reward learning method consists of two parts: we first extract the skill-specific reward function parameterization, $\learnedR(.)$, from LLM using a sequence of guiding prompts, then we design an iterative self-alignment procedure to fit the reward function $\learnedR(.)$ using ranking based preference learning.

\subsubsection{Reward Parameterization Proposal}

\paragraph{Challenge}
We utilize LLM to propose the reward function's parameterization, which includes sensory feature selection and the computation structure. Existing IRL methods either manually select feature sets, learning their linear weights \cite{AndrewNg, maxentirl}, or directly input raw observations into a deep neural network\cite{DeepMaxEnt, GCL, AIRL, fIRL}, hoping the over-parameterized deep neural networks can approximate any forms of reward functions. Either ways, selecting the reward function parameterization remains difficult in reward learning. On one hand, introducing the inductive bias - through pre-selected features and a predetermined reward function class - can make reward learning more efficient. However, these manual efforts lessen the autonomy in reward learning, and the fixed function class restricts reward function's expressiveness. Conversely, learning a reward function without an inductive bias requires more expert demonstrations, which can be both costly and time-intensive to obtain.

\paragraph{Motivation}
The Large Language Model (LLM) contains useful task-specific information, such as identifying crucial attributes necessary for task completion. For example, consider the task of touching: LLM can distinguish from the full set of sensory observations, those subsets that are important for successful touch execution. Features such as \lstinline{obs[`joint_positions']} and  \lstinline{obs[`target_object_pos']} are meaningful because they jointly determine the success of the touch, while irrelevant attributes such as \lstinline{obs[`joint_velocity']} are discarded, as velocity may not be imperative for a simple touch task. Furthermore, the LLM inherently comprehends how these observations should change to achieve task completion and subsequently generates appropriate Python code. 

\paragraph{CoT Reward Parameterization Proposal}
We propose to apply the Chain-of-Thought (CoT) to guide the Large Language Model (LLM) in the generation of reward function parameterization proposal. Our process involves a 4-step procedure, which aids the LLM in deconstructing and comprehending a physical skill, derived from a singular natural language description such as \lstinline{touching}. We further delineate this procedure in subsequent sections. See Figure \ref{fig:para_proposal} for an illustration of our parameterization proposal process.

First, using the full ranges of \lstinline{robot operations} and \lstinline{sensory inputs}, we prompt the LLM to describe the target skill in natural language. This approach helps the LLM relate the physical skill to the context of available operations and sensory feedback. We refrain from immediately asking the LLM to generate code as its training in natural language allows it to provide a more detailed and coherent description of the skill. Essentially, we exploit the LLM's extensive training in natural language to initially describe the skill, before translating it into code.

In the second step, we instruct the LLM to determine the essential set of sensory observation inputs needed for the target skill completion, along with an explanation for this selection. We then ensure that this selection aligns with the broader CoT level, thus maintaining LLM's self-consistency.

In the third step, we instruct the LLM to generate a list of ``Dos'' and ``Don'ts''. The ``Dos'' encompass desirable observations based on sensory data, whereas ``Don'ts'' represent actions or consequences we must avoid. Together, they establish preferences regarding operations and observations crucial for mastering the intended skill. Moreover, by additionally querying the LLM to generate ``Don'ts'', our reward parameterization explicitly encode the undesirable behaviors, allowing it to better distinguish between unwanted and potentially hazardous actions. Unlike traditional reward learning from expert demonstration, which considers all behaviors not part of the expert demonstrations as equally negative, this approach allows not just for differentiation, but for prioritization based on risk levels.

For our final step, we prompt the LLM to develop the Python function \lstinline{get_reward()} based on previously provided responses. The resulting Python code, which can interface with existing APIs and be easily parsed, generates a template akin to a computational graph or function form of the reward function specific to the skill. Although the framework is generally precise, the absolute values or numbers in the initial proposal are not calibrated for realistic robotic systems. This leads us to the next section on refining these values via self-alignment.

\subsection{Updating Reward: Self-Alignment}

\begin{figure}[]
\centering
    \includegraphics[width=1\linewidth]{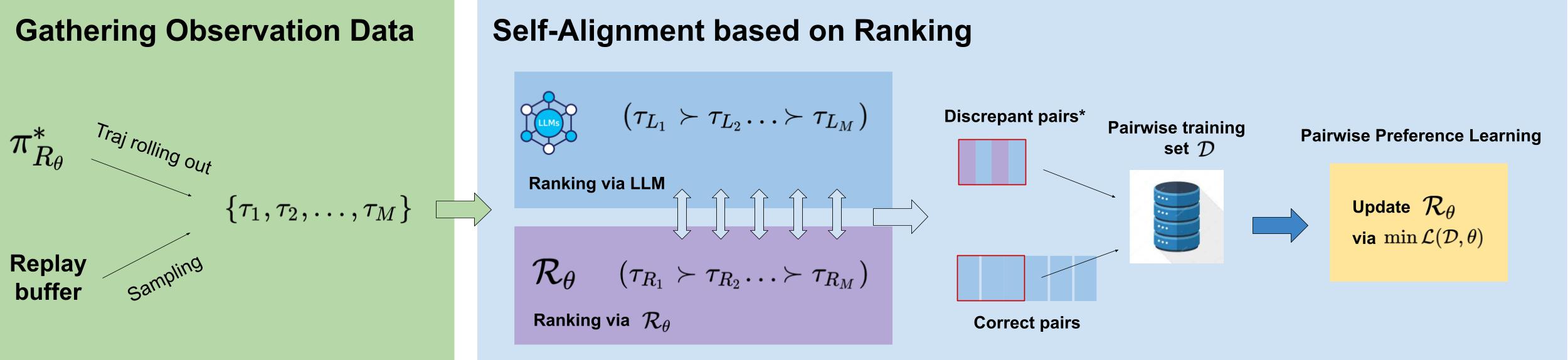} 
\caption{\small \textbf{Illustration for Self-Alignment Reward Update}. For each reward update, we first gather observation data by i) rolling out trajectories using the current policy, and ii) sampling from the replay buffer. Next, we rank these trajectories using LLM and our learned reward function $\learnedR$ respectively. We align the ranking via $\learnedR$ to that of LLM by applying preference learning on all all mismatched pairs and a subset of correctly ranked ones. *When there are no discrepancies, we further query LLM for a failure analysis \cite{liu2023reflect} based on the observation description. This helps to further improve the correctness of the reward function towards successful task completion.}
\label{fig:self-alignment}
\end{figure}

\paragraph{Challenge}
After generating of the reward function template, we learn the numerical values of this parameterization, which ultimately allow the reward function to induce a policy that effectively masters physical skills. To ensure the effectiveness of this policy, the reward function must consider system dynamics and action outcomes. How can we generate these observation signals to supervise the reward function learning with no human-in-the-loop?

\paragraph{Motivation}
Our observations suggest that the LLM can effectively assess the success of the physical skills based on observed signals, making it an effective discriminator for task performance assessment. Inspired by this, we suggest to use LLM's  evaluation to supervise the learning of the reward function. Specifically, to update the reward function, we sample trajectories from recent policy and replay buffer, rank them using both the LLM and the learned reward function respectively, and minimize discrepancies in these rankings. 

\begin{algorithm}[]
\caption{Self-Alignment Reward Update}\label{alg:self-alignment}
\begin{algorithmic}[1]
\Require Learned Reward $\learnedR$, replay buffer $\replay = \varnothing$, initial policy $\pi_{\theta}$
\For {$t = 0, 1, ... $}
    \For {$k = 0, 1, ... $}
    \State Sample transitions using $\pi_{\theta}$ add them to $\replay$
    \State Update $\pi_{\theta}$ using DrQ-v2 with $\learnedR$ and $\replay$
    \EndFor
    \State Sample $M$ trajs $\{\tau_i\}_{i=1,...,M}$using $\pi_{\theta}^*$ 
    \State Sample $N$ trajs $\{\tau_i\}_{i=1,...,N}$ from $\replay$ based on reward histogram
    \State Calculate \rank{\learnedR} by ranking $\{\tau_i\}$ using \learnedR
    \State Calculate \rank{LLM} by ranking $\{\tau_i\}$ using LLM
    \State $\mathcal{D}_{neg} \gets$ discrepancy(\rank{\learnedR}, \rank{LLM})
    \State $\mathcal{D}_{pos} \gets $ $|\mathcal{D}_{neg}|$ pairs sampled from agreed(\rank{\learnedR}, \rank{LLM})
    \State \dataset $\gets \mathcal{D}_{neg}$ + $\mathcal{D}_{pos}$
    \State Update $\theta$ by minimizing $\loss(\theta)$ using \dataset \; according to Equation \ref{equ:loss}.
\EndFor
\end{algorithmic}
\end{algorithm}

\paragraph{Self-Alignment Reward Update} Our double-loop structure reward update, inspired by the bi-level optimization concept in IRL, aims to optimize self-alignment. In particular, the inner loop optimizes the policy based on the current reward function $\learnedR$. While the outer loop updates the reward function $\learnedR$ parameters, aligning the LLM-proposed ranking with that of the learned reward function over $M$ sampled trajectories. Ideally, a successfully learned reward consistently should match its ranking with the LLM. We present the pseudo code in Algorithm \ref{alg:self-alignment}.

We opt for rankings, not absolute scores, to suit the LLM's restricted numerical reasoning capacity. We use the off-policy RL algorithm DrQ-v2\cite{yarats2021mastering} for policy learning here for its effective exploring and exploiting of the environment. We use reward relabeling to reuse the previous experience. More details can be found in appendix. We next focus on elaborating the outer loop's self-alignment update in the subsequent sections.

We suggest a self-alignment process to update the reward function in the outer loop. In each iteration, we first draw $M+N$ trajectories from (1) the replay buffer according to the reward histogram or (2) rollout trajectories from the latest policy. We then retrieve the two ranking sets from the present reward function $\learnedR$ and by querying from LLM. We then align the ranking from $\learnedR$ with that of LLM using pairwise preference learning. To generate a dataset $\dataset$ of pairwise comparison, $\dataset = \{ (\traj{a}^0, \traj{b}^0)_0, (\traj{a}^1, \traj{b}^1)_1, \ldots\}$ where $\traj{a}^i \succ \traj{b}^i$, we sampled paired trajectories $\traj{i} \succ \traj{j}$ from LLM ranking. In practice, we require LLM to cluster the trajectories with similar performance to prevent comparing similar or incomparable states. The ranking is done over data from different clusters only. Thus, the enhanced pairwise dataset emphasizes significant performance discrepancies through a partial order. We then select all pairs where the ranking based on $\learnedR$ disagrees with the LLM ranking and add them to the dataset $\dataset$. To maintain the progress already achieved by the reward function, we sample equal amounts of correctly ranked pairs and add them into $\dataset$ and randomly shuffle it. 

Similar to modeling noisy optimal human behavior with Boltzmann rationality, we represent LLM with a Boltzmann-rational model, which assumes it is more likely to choose a higher reward item if it is more rational (i.e. a higher $\beta$). This noisy optimal behavior model aligns with our motivation, without requiring perfect ranking feedback but only the majority of them. We found it to be the case as LLM occasionally mis-ranks with subtle numerical differences (i.e. 0.001 scale). Under $\learnedR$, the preference of one trajectory over another can be modeled as a softmax-normalized distribution:

\begin{equation}
     \hat{P}[\traj{i} \succ \traj{j}] = \frac{\exp\beta(\underset{s^i\in \traj{i}}{\sum}\learnedR(s^i))}{\exp\beta(\underset{s^i\in \traj{i}}{\sum}\learnedR(s^i)) + \exp\beta(\underset{s^j\in \traj{j}}{\sum}\learnedR(s^j))}.
\end{equation}

With the pairwise comparison feedback, the reward model is updated using Bayesian inference. As the reward function parameter $\theta$ is continuous which leads to an intractable normalizer, we follow previous works and use Metropolis-Hastings algorithm to sample $\theta$ from the unnormalized posterior. The parameters are updated iteratively to achieve lower cross-entropy differences between these predictions based on the learned reward function and the labels in the feedback. The updated parameter will be accepted if and only if such inconsistency reduces and converges. 

\begin{align}
    \loss(\theta) =  - \sum_{(\traj{i}, \traj{j}) \in \dataset} \log  \hat{P}[\traj{i} \succ \traj{j}]  = - \sum_{(\traj{i}, \traj{j}) \in \dataset} \frac{\exp\beta(\underset{s^i\in \traj{i}}{\sum}\learnedR(s^i))}{\exp\beta(\underset{s^i\in \traj{i}}{\sum}\learnedR(s^i)) + \exp\beta(\underset{s^j\in \traj{j}}{\sum}\learnedR(s^j))}
    \label{equ:loss}
\end{align}

When there is not discrepancy, yet no successful trajectory is identified from LLM, we further prompt the LLM to analyze the blocking factor and its relevant reward or penalty term based on the data sample, then update the parameter based on this failure analysis\cite{liu2023reflect}. This failure analysis guides our parameter tuning of the reward function to strengthen the identified reward or penalty signal. An example of this is included in Appendix C.
\section{Experimental Results}
\label{sec:experiment}

\begin{table}[]
    \centering
    \begin{tabular}{ccc}
    \hspace{-5mm}
         \includegraphics[width=0.32\textwidth]{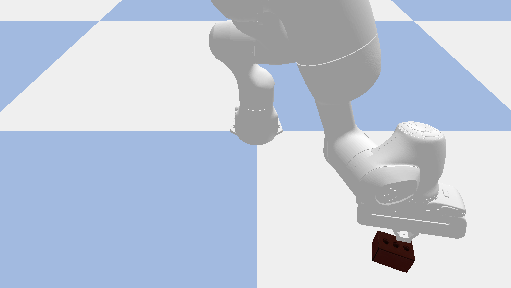}
         &
        \includegraphics[width=0.32\textwidth]{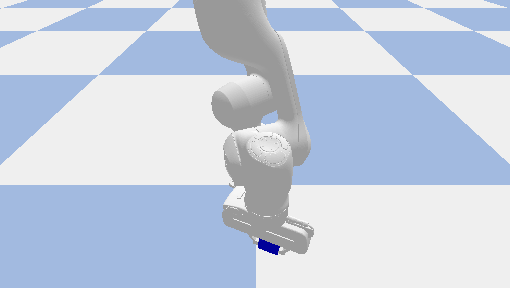}
         &
        \includegraphics[width=0.32\textwidth]{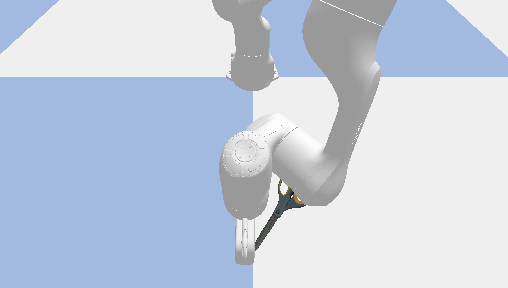}
    \end{tabular}
    \caption{\textbf{Task illustration: (l) touching; (m) grasping and lifting, (r) pushing to target position.} 
    }
    \label{tab:tasks}
\end{table}

\subsection{Overview}
We conducted simulated tests focusing on three physical skills, touching, grasping and lifting, pushing to the target position. Our experimental design was to objectively evaluate three main hypotheses:
\begin{enumerate}[label={\textit{\textbf{H\protect\digits{\theenumi}}.}}]
    \item The reward function we developed can establish the optimal policy faster than the ground-truth sparse reward function. 
     \item The reward function proposed by LLM, when relying solely on CoT, cannot consistently generate a policy capable of mastering the physical skills.
    \item By utilizing self-alignment, we can learn the parameters of LLM's proposed reward function, allowing the learned reward to guide a policy to successfully master the physical skills.
\end{enumerate}

We compare our approach against two baselines. The first baseline exploits the ground-truth sparse reward function from the simulator. We substantiate \hypo{1} by evaluating its training performance over a fixed number of training steps and the total steps required to reach the stable state. The second baseline uses CoT to propose reward function parameterization but does not apply self-alignment to change the parameters of the proposed reward function. The final performance of this second standard approach was used as evidence for \hypo{2}. Moreover, as the key distinction between our approach and the second baseline lies in action grounding, any performance discrepancy between the two serves as evidence to validate \hypo{3}. 

\subsection{Environment Setup}
We evaluated the pipeline in 3 physical skills learning in a tabletop manner in PyBullet simulation with a fixed horizon of 25 steps. A Cartesian impedance controller is implemented to enable compliant control. The action space includes the Cartesian position and yaw changes as $a = (\delta_x, \delta_y, \delta_z, \delta_\theta)$.

\textbf{Touching} Starting from the home position, the robot is expected to reach a goal object spawn randomly and maintain contact with it. Selected by LLM, the state includes robot joint positions, the end effector pose, the end effector contact force, and the target object position. 

\textbf{Grasping and Lifting} Similarly, the robot is expected to grasp and lift up a cube spawn randomly in front of it. The grasping action is automatically applied when an external force is detected from the end-effector. Selected by LLM, the state includes robot joint positions, the end effector pose, the end effector contact force, and the target object position. 

\textbf{Pushing to Goal Position} For this task, the robot is expected to push a scissor spawned randomly to a target position. This task is more challenging for two reasons (1) The policy involves learning the dynamics to enable accurate pushing especially given the scissor has an unevenly distributed mass; (2) The object is slim that the robot needs to carefully move to produce a push while avoiding collision. The state for this task includes robot joint positions, the end effector pose, the end effector contact force, the target object position, and the goal position; 
 
We use DrQ-v2 \cite{yarats2021mastering} as an off-policy method to learn the control policy. The details for hyper-parameters are in Appendix E. The reward function is updated on every 1e4 step. For the Boltzmann-rational model, we set $\beta$=0.9. For reward updates from preference ranking, we used a customized implementation to accept flexible reward forms and parameters based on APReL \cite{biyik2022aprel}. 

\subsection{Result Analysis}
\begin{figure}[]
\centering
    \includegraphics[width=1\linewidth]{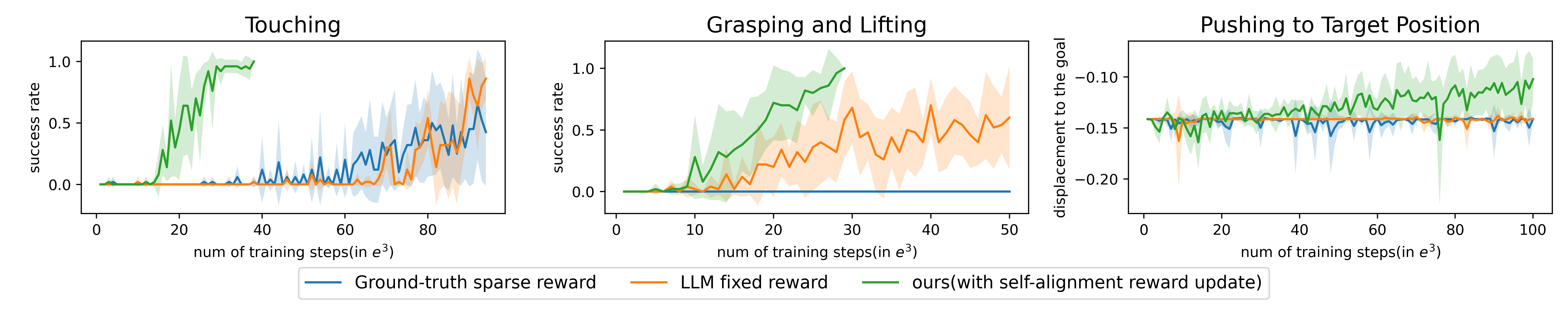} 
\caption{\small \textbf{Training curves for our method and the two baselines}. We measure performance in touching, grasping-and-lifting using success rates, and pushing tasks using final negative distance to the goal. Higher measurements indicate better performance in all three tasks.}
\label{fig:training_curves}
\end{figure}

We compare the training curves and the performance evaluation for using sparse reward, fixed LLM reward, and updated LLM reward with self-alignment. For each task and method, the experiments run over 5 different seeds. An early stop is set when the model reaches 100\% success rates as the reward update stops. The results are summarized in Fig \ref{fig:training_curves}. Next, we analyze the method performance from the effectiveness and efficiency perspectives.

\textbf{Effectiveness.} With self-alignment to update the reward function, the learned reward function's parameters are numerically stable and can induce policy to master the given task, while sparse reward fails for grasping and lifting tasks. Both sparse reward and fixed LLM reward failed to learn pushing action. We think the effectiveness of self-alignment learned reward function comes from two aspects: 

(1) LLM brings meaningful inductive bias and decomposes task objectives into dense reward and penalty terms. Such multi-objective learning especially for longer horizon tasks can be challenging to learn altogether. By feedback on recent exploratory behaviors from the replay and exploitative behavior from current policy to LLM, LLM identifies the most and least optimal behaviors under the local context and steers it towards the global goal. Such behavior is similar to introducing a curriculum for learning with periodic feedback and can be observed from the weight parameter change. For example, for pushing to target position task (details in appendix F table \ref{table:reward_upate_example}), reward transits from encouraging approaching first, later when contact starts it with policy improving, the weights further shift to encourage pushing and maintaining to target position. By iteratively doing so, the final policy was effectively guided toward the final target step by step. 

(2) As addressed in our motivation, LLM is more unreliable when generating numerical output, which is crucial to a reward function design. This can be observed by a slow converging rate or failing to converge in the fixed LLM training curve. Meanwhile, reward shaping is nontrivial as inappropriate shaping will introduce undesired inductive bias that leads to sub-optimal behavior of policy by exploiting such flaws. Through a closed-loop reward update via self-alignment, such numerical errors or flaws can be reflected through feedback and corrected via ranking. For example, in the pushing task, when given a high reach reward and low push reward, the policy converges to touch the scissor, as the initial push action easily loses contact with the object and yields a lower reward. However, LLM ranks with a preference for a nearer distance between the object and the goal (details in Appendix D). The reward is updated to increase the push weight from 2.0 to 21.05 iteratively to correct such sub-optimality and induce more pushing actions. Another example is on touching task, LLM initially proposed a gentle touch force to range from 1 to 5N. After observing multiple data samples with contact force $\leq$ 1N, LLM clusters those samples as successful touching and enables the minimum force threshold to update to 0.127N.

\textbf{Efficiency.} From figure \ref{fig:training_curves}, the learned reward successfully trained target behavior with much fewer training steps than the baselines across all tasks. For touching task. it takes an average $24,600$ steps for the policy to reach 100\% success rate. For a fixed LLM reward, it takes $93,200$ steps. For sparse reward, only 4 out of 5 runs reach 100\% success rate within 100k steps and take on average $98,000$ steps among them. Similarly, for grasping and lifting task that self-alignment updated rewards takes $19,000$ steps to reach 100\% success rate, while only 3 out of 5 LLM fixed reward runs reach 100\% success rate within 50k steps and take on average $45,000$ steps among them. No policies trained using sparse rewards learn to grasp and lift successfully within the same steps. 

\section{Conclusion}
In this work, we propose a self-alignment method that learns the reward function of physical skills from LLM without the need for expert demonstration or feedback. This method efficiently learns precise control and fine-grained sensory input. We assessed our method on three simulated tasks. The results show that our reward function encourages optimal behaviors, requiring significantly less training than existing baseline methods. The limitation of this work includes when a compilation error is encountered in the initial reward function, it requires one-time human intervention to correct it and we are looking into more automatic self-correction for this.

\bibliography{example}  
\newpage
\appendix
\section{Full Prompts}
\subsection{Background: Robot, Sensory API}
The exact importing and function implementation is omitted. Only function structure and docstrings are prompted to the LLM. The LLM is required to summarize available methods into a dictionary of available with input and output explanation.

\subsection{Task Prompt}
\begin{lstlisting}
    Our goal is to learn a set of sensory-motor skills from language concepts with multi-modal sensory observations with robots interacting with the environment and learn through trial and error in the reinforcement learning manner. We have haptic and proprioceptive sensors equipped to the environment. In the environment, we have: 
        (1) a robot (access by env.robot) which we can communicate and control using the robot interface as showed previously; 
        (2) a target object (env.target_object) and a list of other objects (env.objects) present in the environment usually the motion should avoid to collide with. The number of other objects varies and can be 0. All objects are of type RigidObject where we interact using the rigid object interface. 
        (3) the goal position (env.goal_pos) as the x, y, z world Cartesian coordinate of the goal position if there is. 
        
        
        The available observations from the environment are stored in a dictionary obs which includes: 
        1. obs['joint_positions']: robot joint positions; 
        2. obs['joint_velocities']: robot joint velocities; 
        3. obs['joint_reaction_forces']: robot joint reaction forces; 
        4. obs['ee_pose']: end effector pose; 
        5. obs['ee_contact_force']: the external contact force the robot end effector is experiencing; 
        6. obs['target_object_pos']: target object centroid position (if there is a target object specified in the task). 
        7. obs['object_poses']: other object centroid positions. 
        
        The robot is supposed to learn a sensorimotor policy with a neural network using reinforcement learning. When given observation as input the policy outputs the end effector pose action. The goal of the policy is to learn optimal action decision-making for a given skill. The robot is controlled with delta end effector pose command, that is every time the end effector is commanded to move d_x, d_y, d_z, d_roll, d_pitch, d_yaw. All skills are assumed to start from a pre-manipulation stage the gripper is about 15cm away from the target to be manipulated, therefore we only learn the end-to-end policy, and no approaching and motion planning is required. 
        
        To start, the first step is to pick the relevant observations from 1-9 observations described above as the policy input when given a skill to learn. The selection should select the most informative and relevant observations as though more observations may provide more information, it requires a large policy network to process it and requires more training data and time which is strongly discouraged here. Please do that for the grasping and lift an object above the ground skill. The robot is expected to grasp a target object to the goal x, y, z position and maintain it there. The grasping action is automatically applied if it is in contact with the target object, and open the gripper otherwise. Please organize your answer into four parts: 
        1) describe the skill; 
        2) analyse if this skill requires gripper control. Use 1 for yes and 0 for no. Should the gripper start with open or closed fingers? Use 1 for open and 0 for closed.
        3) select the relevant and minimal set of observations. 
        
        Lastly, summarize the answer for 2 and 3 into a separate paragraph. Please only include the answer 1/0, 1/0 only for question 2, and a list of observation indices for question 3. For example: "1, 1, [1, 2, 3, 4, 5]"

\end{lstlisting}

\subsection{Reward Function Prompt}
\begin{lstlisting}
    Now given the observation of the current step, how would you define the reward function env.get_reward(obs) for learning <skill>? The target object can be of arbitrary shape. The collision can be checked with env.detect_collision() and should always be penalized. 

    Please: 
    1) analysis of steps to complement the task. Identify all relevant signals and assign rewards to encourage them to happen until the task is completed. During the process, what are the don'ts? assign penalties to prevent them from happening. For each item, are there any properties or methods in the robot/object interfaces relevant to it? 
    2) for all rewards and penalties, think about how should it be calculated. Is it a continuous or constant value under certain conditions? Use a continuous implementation if you can. 
    3) with analysis from 2), implement the final get_reward() function. Add comments explaining the chain of thought on top of each line of code. 
    
    For configurable hyper-parameter, please put all value settings as the function input arguments. Never directly assign a magic value to a parameter in the function. The correct and wrong examples are shown below
    [Correct]: 
    def get_reward(obs, parameter_name=1.0) 
    [Wrong]: def get_reward(obs): parameter_name=1.0 
    
    Lastly, please identify the hyper-parameter range and output as a dictionary, the key is the hyper-parameter name, and the value is a list of possible minimum and maximum parameter values of this parameter. Use np.inf for positive infinite and -np.inf to represent negative infinite value and make sure the range is indeed the largest range of the parameter.

\end{lstlisting}

\subsection{Ranking Prompt}
\begin{lstlisting}
    The reward function you provided is: 
    <function definition>
    
    Given execution observation for 
    - data sample 1: <obs_name>=<obs_value>
    - data sample 2: <obs_name>=<obs_value>
    - data sample 3: <obs_name>=<obs_value>
    - data sample 4: <obs_name>=<obs_value>
    - data sample 5: <obs_name>=<obs_value>
    - data sample 6: <obs_name>=<obs_value>
    - data sample 7: <obs_name>=<obs_value>
    - data sample 8: <obs_name>=<obs_value>
    - data sample 9: <obs_name>=<obs_value>
    - data sample 10: <obs_name>=<obs_value>
    
    For all the samples, please identify the samples with successful pushing the target object to the target position execution. Among all the unsuccessful samples, please cluster data so similar states will not be ranked.  For each cluster use one data sample and provide the rank among all clusters only if they are significantly better.
    Please do not output anything after the two lists. For example:
    Result:
    [1, 3], [5, 2, 4]

\end{lstlisting}

\subsection{Hyper-parameter Tuning Prompt When No Discrepency}
\begin{lstlisting}
    Consistent ranking found.
    From the observation, go through each data sample and check if it succeeds in executing <skill>. What is the top factor that blocks it from a better execution and what reward or penalty terms relate to this factor?
    After going through all samples, count the times that each relevant reward or penalty term is mentioned. Please output the identified hyper-paramter as a dictionary. The key is the hyper-parameter name and value is the recommended new value. such as: 
    {'param_a': 1.0, 'param_b': 1.0}
\end{lstlisting}

\section{Reward Function}
\subsection{Touching}
\textbf{Sparse Reward Function}
\begin{lstlisting}
    def get_reward(obs, force_threshold=5):
        if self.robot.gripper_in_contact(self.target_object):
            total_reward = 1
        else:
            total_reward = 0
            
        return total_reward
\end{lstlisting}

\textbf{LLM Generated Reward Function}
\begin{lstlisting}
    def get_reward(obs, distance_weight=1.0, contact_bonus=10.0, min_contact_force=1.0, force_threshold=5.0, collision_penalty=-10.0):
        # Calculate the Euclidean distance between the end effector and the target object
        distance = np.linalg.norm(obs["ee_pose"][:3] - obs["target_object_pos"])
        # Negative reward based on distance to the target
        distance_reward = -distance_weight * distance
    
        # Check the magnitude of the force applied by the end effector
        force_magnitude = np.linalg.norm(obs["ee_reaction_forece"])
        # Provide a reward if a meaningful force is detected (indicating a touch)
        contact_reward = contact_bonus if force_magnitude > min_contact_force else 0
    
        # Penalize if the force applied by the end effector is too high
        force_penalty = -force_threshold if force_magnitude > force_threshold else 0
    
        # Check for collisions with other objects
        collision_detected = any([obj.in_contact(env.robot) for obj in env.objects])
        # Penalize if any collision is detected
        collision_penalty_value = collision_penalty if collision_detected else 0
    
        # Calculate the total reward
        total_reward = distance_reward + contact_reward + force_penalty + collision_penalty_value
    
        return total_reward
\end{lstlisting}
\subsection{Grasping and Lifting}
\textbf{Sparse Reward}
\begin{lstlisting}
    def get_reward(obs, **kwargs):
        if env.robot.gripper_in_contact(env.target_object) and (not env.target_object.in_contact(env.floor)):
            total_reward = 1.0
        else:
            total_reward = 0.0
            
        return total_reward

\end{lstlisting}
\textbf{LLM Generated Reward Function}
\begin{lstlisting}
    def get_reward(obs, distance_weight=1.0, contact_bonus=5.0, height_weight=2.0, collision_penalty=-10.0):
        # Distance to target
        distance_to_target = np.linalg.norm(np.array(obs['ee_pose'][:3]) - np.array(obs['target_object_pos']))
        reward_distance = -distance_weight * distance_to_target  # Closer the distance, higher the reward
    
        # Check if gripper is in contact with the target object
        contacted = env.robot.gripper_in_contact(env.target_object)
        reward_contact = contact_bonus if contacted else 0.0
    
        # Height reward
        object_height = obs['target_object_pos'][2]
        reward_height = height_weight *  max(0, object_height - 0.01)
    
        # Collision check
        collision_detected = env.detect_collision()
        reward_collision = collision_penalty if collision_detected  else 0.0
    
        total_reward = reward_distance + reward_contact + reward_height + reward_collision
        
        return total_reward
\end{lstlisting}

\subsection{Pushing To A Target Position}
\textbf{Sparse Reward Function}
\begin{lstlisting}
    def get_reward(obs, threshold=0.025):
        distance = np.linalg.norm(np.array(obs["target_object_pos"]) - np.array(env.goal_pos))
        total_reward = 1.0 if distance <= threshold else 0.0
        
        return total_reward
\end{lstlisting}
\textbf{LLM Generated Reward Function}
\begin{lstlisting}
    def get_reward(obs, reach_weight=0.5, push_weight=2.0, maintain_weight=5.0, collision_penalty=-10.0):
        # Distance between end effector and target object
        distance_to_target = np.linalg.norm(np.array(obs['ee_pose'][:3]) - np.array(obs['target_object_pos']))
    
        # Distance between target object and goal
        distance_to_goal = np.linalg.norm(np.array(obs['target_object_pos']) - np.array(env.goal_pos))
    
        # Reward for reaching near the object
        reach_reward = - reach_weight * distance_to_target
    
        # Reward for pushing object towards goal
        push_reward = - push_weight * distance_to_goal
    
        # Reward for maintaining object at goal
        maintain_reward = maintain_weight if distance_to_goal < 0.05 else 0
    
        # Check for collisions
        collision_detected = env.detect_collision()
        collision_penalty_val = collision_penalty if collision_detected else 0
    
        # Total reward
        total_reward = reach_reward + push_reward + maintain_reward + collision_penalty_val
    
        return total_reward

\end{lstlisting}

\section{Failure Analysis Example}
\textbf{Data Sample}
\begin{lstlisting}
  - data sample 0: distance_to_target = 0.2471, contacted = False, object_height = 0.0250, collision_detected = False.
  - data sample 1: distance_to_target = 0.2811, contacted = False, object_height = 0.0250, collision_detected = False.
  - data sample 2: distance_to_target = 0.2420, contacted = False, object_height = 0.0250, collision_detected = False.
  - data sample 3: distance_to_target = 0.4353, contacted = False, object_height = 0.0250, collision_detected = False.
  - data sample 4: distance_to_target = 0.4579, contacted = False, object_height = 0.0250, collision_detected = False.
  - data sample 5: distance_to_target = 0.1746, contacted = False, object_height = 0.0250, collision_detected = False.
  - data sample 6: distance_to_target = 0.2889, contacted = False, object_height = 0.0250, collision_detected = False.
  - data sample 7: distance_to_target = 0.3366, contacted = False, object_height = 0.0250, collision_detected = False.
  - data sample 8: distance_to_target = 0.4867, contacted = False, object_height = 0.0250, collision_detected = False.
  - data sample 9: distance_to_target = 0.4483, contacted = False, object_height = 0.0250, collision_detected = False.
  
   (ranking and checking. No discrepancy is found)

    From the observation, go through each data sample and check if it succeeds in executing grasping and lifting the target object to the target position. What is the top factor that blocks it from a better execution and what reward or penalty terms relate to this factor?
    After going through all samples, count the times that each relevant reward or penalty term is mentioned. Please output the identified hyper-paramter as a dictionary and do not include other hyper-parameters. The key is the hyper-parameter name and value is the recommended new value. such as: 
    {'param_a': 1.0, 'param_b': 1.0}
\end{lstlisting}
\textbf{Failure Analysis}
\begin{lstlisting}
    Analysis of Each Sample:
    
    Sample 0:
    
    Distance to target: 0.2471
    Gripper Contact: No
    Object Height: 0.0250
    Collision: No
    Analysis: The robot did not make contact with the object, and the object hasn't been lifted (since the height remains close to the floor height). The primary reason here is the distance to the target.
    Relevant Terms: distance_weight
    
    Sample 1:
    
    Distance to target: 0.2811
    Gripper Contact: No
    Object Height: 0.0250
    Collision: No
    Analysis: Same as sample 0. The robot didn't manage to get close enough to the target.
    Relevant Terms: distance_weight
    
    Sample 2:
    
    Distance to target: 0.2420
    Gripper Contact: No
    Object Height: 0.0250
    Collision: No
    Analysis: Very similar to sample 0 and 1. Distance remains the issue.
    Relevant Terms: distance_weight
    
    ... (and so on for all the other samples) ...
    
    Given that all samples share a similar pattern and do not make contact with the object, and the object height is consistently low (close to floor height), the primary issues can be summarized as:
    
    Distance to target: The robot isn't getting close enough to the object to initiate a grasp. The related term is distance_weight.
    
    No gripper contact: The robot isn't making contact with the object. The reward for making contact is contact_bonus. The lack of a positive reward here might imply that the robot isn't finding it beneficial to make the contact.
    
    Low object height: The object remains close to the floor in all samples, indicating unsuccessful lifting. The relevant term is height_weight.
    
    Count of times each relevant term is mentioned:
    
    distance_weight: 10
    contact_bonus: 10
    height_weight: 10
    Given that all samples have the same issues, the count for each term remains consistent.
    
    Recommended Hyper-parameter Adjustments:
    To encourage the robot to approach the object and make contact, and subsequently lift it, we might want to:
    
    Increase distance_weight to provide a stronger reward signal as the robot gets closer to the target.
    Increase contact_bonus to give the robot a more significant reward when making contact.
    Increase height_weight to further incentivize the robot to lift the object higher.
    Output:
    {'distance_weight': 2.0, 'contact_bonus': 10.0, 'height_weight': 4.0}
\end{lstlisting}

\section{Ranking Example}
\textbf{Feedback with Sample Observation Description}
\begin{lstlisting}
    The reward function you provided is:                                                     
    
    def get_reward(obs, reach_weight=9.320304774411772, push_weight=3.1609988615863727, maintain_weight=5.082954771564665, collision_penalty=-10.709883122410613):
        # Distance between end effector and target object
        distance_to_target = np.linalg.norm(np.array(obs['ee_pose'][:3]) - np.array(obs['target_object_pos']))
    
        # Distance between target object and goal
        distance_to_goal = np.linalg.norm(np.array(obs['target_object_pos']) - np.array(env.goal_pos))
    
        # Reward for reaching near the object
        reach_reward = - reach_weight * distance_to_target
    
        # Reward for pushing object towards goal
        push_reward = - push_weight * distance_to_goal
    
        # Reward for maintaining object at goal
        maintain_reward = maintain_weight if distance_to_goal < 0.05 else 0
    
        # Check for collisions
        collision_detected = env.detect_collision()
        collision_penalty_val = collision_penalty if collision_detected else 0
    
        # Total reward
        total_reward = reach_reward + push_reward + maintain_reward + collision_penalty_val
    
        return total_reward
    
    Given execution observation for
      - data sample 0: distance_to_target = 0.0235, distance_to_goal = 0.1833, collision_detected = False.
      - data sample 1: distance_to_target = 0.0257, distance_to_goal = 0.1401, collision_detected = False.
      - data sample 2: distance_to_target = 0.0232, distance_to_goal = 0.1538, collision_detected = False.
      - data sample 3: distance_to_target = 0.0266, distance_to_goal = 0.1531, collision_detected = False.
      - data sample 4: distance_to_target = 0.0234, distance_to_goal = 0.1258, collision_detected = False.
    
    For all the samples, please identify the samples with successful pushing the target object to the target position execution. Among all the unsuccessful samples, please cluster data so similar states will not be ranked.  For each cluster use one data sample and provide the rank among all clusters only if they are significantly better.
    
    Please do not output anything after the two lists. For example:
    Result:
    [1, 3], [5, 2, 4]
\end{lstlisting}
\textbf{LLM Clustering and Ranking}
\begin{lstlisting}
     
    Analysis:
    
    Successful Samples:
    To be successful, the distance_to_goal should be less than 0.05 meters. However, none of the samples meet this criterion, so there are no successful samples.
    
    Unsuccessful Samples:
    To cluster the data based on similar states, we'll consider the following features:

    Object Proximity to the Gripper (distance_to_target): Smaller values suggest the robot's gripper is closer to the object, which is a positive indication for a potential push.
    Distance to Goal (distance_to_goal): A smaller distance indicates that the object is closer to the target position.
    Given that all the samples have collision_detected = False, collision does not factor into our clustering.
    
    Let's analyze the clustering:

    Cluster 1:
    Sample 4 with distance_to_goal = 0.1258

    Cluster 2:
    Sample 1 with distance_to_goal = 0.1401
    Sample 2 with distance_to_goal = 0.1538
    Sample 3 with distance_to_goal = 0.1531
    From Cluster 2, Sample 1 can be chosen to represent the cluster because of its smallest distance_to_goal in comparison with other samples in the same cluster.
    
    Cluster 3:
    Sample 0 with distance_to_goal = 0.1833

    Ranking clusters based on better performance (with a preference for closer distance_to_goal):
    
    Cluster 1 (Sample 4): This cluster is ranked the best among the unsuccessful samples because it has the smallest distance_to_goal.
    Cluster 2 (Sample 1): It's ranked second as the distance_to_goal for its representative sample is larger than Cluster 1 but smaller than Cluster 3.
    Cluster 3 (Sample 0): This cluster is ranked last among the unsuccessful samples because it has the largest distance_to_goal.
    Output:
    Result:
    [], [4, 1, 0]
\end{lstlisting}

\section{Hyper-parameter Details}
In this section, we provide the hyper-parameter details of the DrQ-v2 \cite{yarats2021mastering} that are used for the three tasks training. The actor and critic network share the same state encoder, where the numerical features are passed to single layer fully connected features and concatenated together as a latent state feature. The latent state feature is further passed to two fully connected layers with 1024 hidden dimensions each followed by a ReLU activation for both actor and critic networks:
\begin{table}[h!]
\begin{center}
\begin{tabular}{l l}
\hline
Hyper-parameter & Value \\
\hline 
lr & 1e-4 \\ 
discount factor & 0.99 \\
batch size & 32 \\
noise stddev min & 0.3 \\
noise stddev max & 1.0 \\
noise stddev decay steps & 10000 \\
noise stddev clip & 0.5 \\
number of exploration steps & 1000 \\
numerical feature dimension & 50 \\
update actor every steps & 2 \\
update critic target every steps & 100\\
critic target tau & 0.01 \\
gradient clip & 20 \\
replay buffer size & 1e6 \\
\hline
\end{tabular}
\end{center}
\caption{DrQ-v2 Hyperparameters.}
\label{table:hps}
\end{table}

\section{Reward Update}
To update the reward function, we train a policy and periodically sample from the recent replay buffer. The sampling was done by analyzing the histogram of the reward distribution in 10 bins from the replay buffer and randomly retrieving one data sample from each non-empty bin. This is to ensure maximum coverage of the data varieties when feedback to LLM. Meanwhile, the latest policy rolls out 5 trajectories to provide data that represents the recent behaviors. Such periodic feedback is done for every 10,000 steps. 

With the updated reward function, the policy continues from the previous weight with reward relabeling. For more flexible reward relabeling, we extract all the reward parameters in the function in Python automatically and store them in the replay buffer instead of the reward value. When sampling from the replay buffer, we reuse all data stored in the replay and a reward will be calculated with the latest $R_\theta$ for policy training. The extracted reward parameters are the same set of variables used to generate the data sample observation for LLM feedback.

During the initial reward function proposal, the LLM is prompted to generate the maximum value range for each parameter. For example: 
\begin{lstlisting}
        hyper_parameter_ranges = {
            "reach_weight": [0.1, np.inf],
            "push_weight": [0.1, np.inf],  
            "maintain_weight": [0.1, np.inf], 
            "collision_penalty": [-np.inf, -0.1]
            }
\end{lstlisting}

To update the weight, it is important to ensure the reward value transits stably. We defined additional regularization terms $\omega$ to shrink the original value space $[V_{min}, V_{max}]$ to:
\[ [max(V_{min}, V_{curr}-\omega), min(V_{max}, V_{curr} + \omega)]  \]
The $\omega$ is defined in a grid of [1, 3, 5, 10] in our experiments and will be conducted for preference fitting in parallel. The updated parameters are selected with maximum discrepancy reduction. When multiple grids achieve the same discrepancy reduction, we select the value that is nearest to the current value. In the case of no discrepancy and hyper-parameter tuning with LLM hinted direction, we choose the one that achieves the maximum performance improvement.

The example of the reward update over iterations is shown here for the pushing to the target position task. We can observe a transition that encourages approaching to pushing and maintaining while avoiding a collision when contact starts with policy improving and behaving more optimally. We can also observe that once the reward logic and policy are coherent, the reward update becomes subtle and behaves similarly to keep a fixed reward design:
\begin{table}[h!]
\begin{center}
\begin{tabular}{c c c c c c}
\hline
 iter & update type & reach weight & push weight & maintain weight & collision penalty\\ 
 \hline
 1 & init & 0.5 & 2.0 & 5.0  & -10.0 \\  
 \hline
 2 & tune(+) reach weight & 10.5 & 2.0 & 5.0  & -10.0  \\  
  \hline
 3 & Bayesian update & 9.32 & 3.16 & 5.08 & -10.71 \\
 \hline
 4 & Bayesian update & 2.05 & 12.72 & 8.01 & -14.11 \\
 \hline
 5 & Bayesian update & 2.59 & 20.92 & 9.22 & -15.99 \\
  \hline
 6 & Bayesian update & 2.54 & 20.93 & 9.23 & -15.94 \\
  \hline
 7 & Bayesian update & 2.56 & 21.19 & 9.24 & -15.95 \\
  \hline
 8 & Bayesian update & 2.53 & 20.94 & 9.40 & -16.07 \\
  \hline
 9 & Bayesian update & 2.47 & 21.02 & 9.10 & -15.98 \\
  \hline
 10 & Bayesian update & 2.48 & 21.05 & 9.31 & -15.98 \\
  \hline
\end{tabular}
\end{center}
\caption{Reward Update over Iteration.}
\label{table:reward_upate_example}
\end{table}

The visualization of the above policy execution over different iterations are illustrated. Five evenly sampled frames including the first and last frame of an episode are selected and concatenated for the transition visualization as explained above:
\begin{figure}[]
\centering
    \includegraphics[width=1\linewidth]{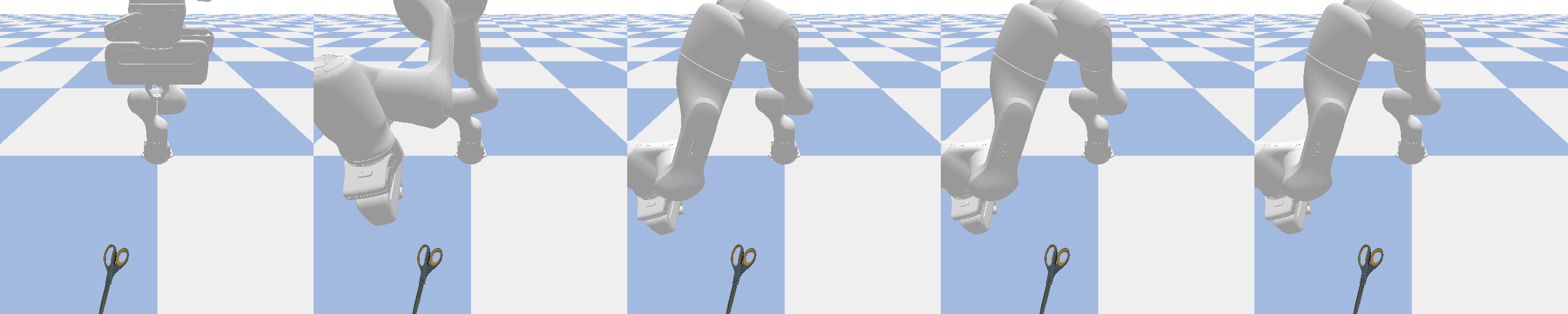} 
    \vspace{-20pt}
\caption{\small Initial reward iteration: policy fails to reach to the target object due to large potential collision region}
\label{fig:iter_1}
\end{figure}
\begin{figure}[]
\centering
    \includegraphics[width=1\linewidth]{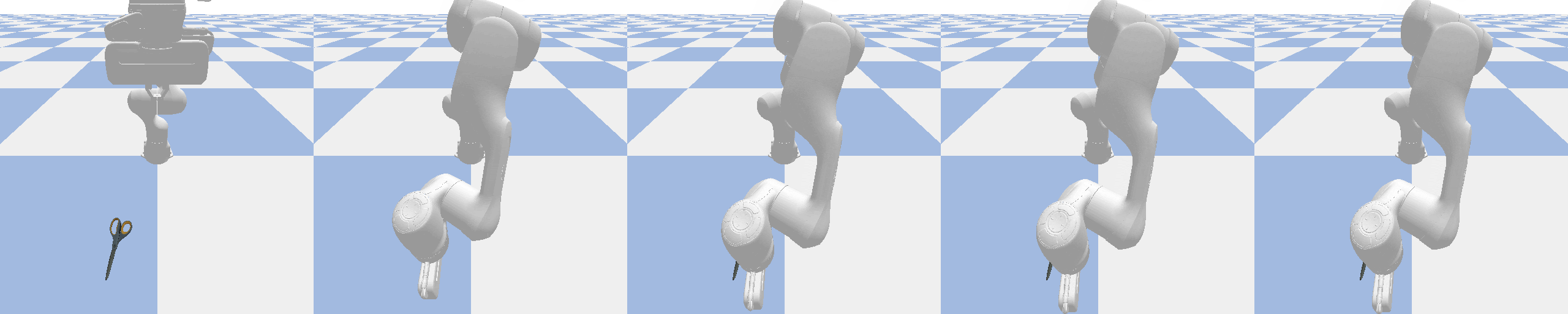} 
    \vspace{-20pt}
\caption{\small Reward update iteration 2: with increased reach weight, the policy starts to reach nearer to the object}
\label{fig:iter_2}
\end{figure}
\begin{figure}[]
\centering
    \includegraphics[width=1\linewidth]{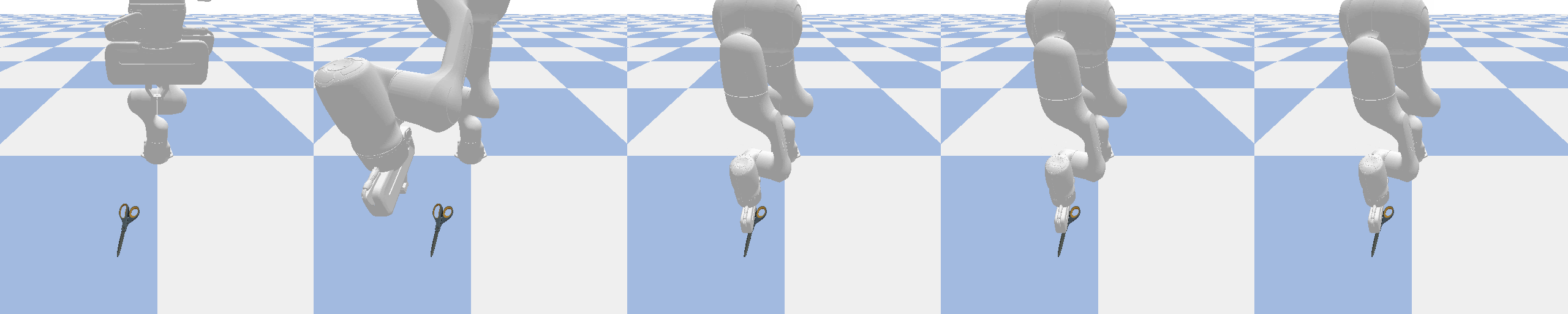} 
    \vspace{-20pt}
\caption{\small Reward update iteration 3: policy starts to make contact with object in all directions. With a relatively high reach reward and low push reward and skill mastered, policy converges to stay touching with the object.}
\label{fig:iter_3}
\end{figure}
\begin{figure}[]
\centering
    \includegraphics[width=1\linewidth]{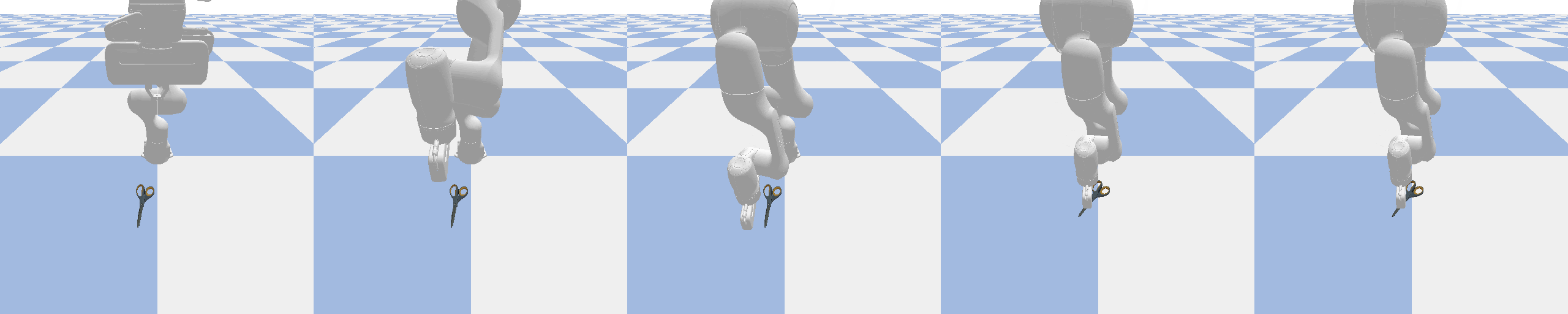} 
    \vspace{-20pt}
\caption{\small Reward update iteration 4: with increased push weight, policy starts to learn to push in the target directions}
\label{fig:iter_4}
\end{figure}
\begin{figure}[]
\centering
    \includegraphics[width=1\linewidth]{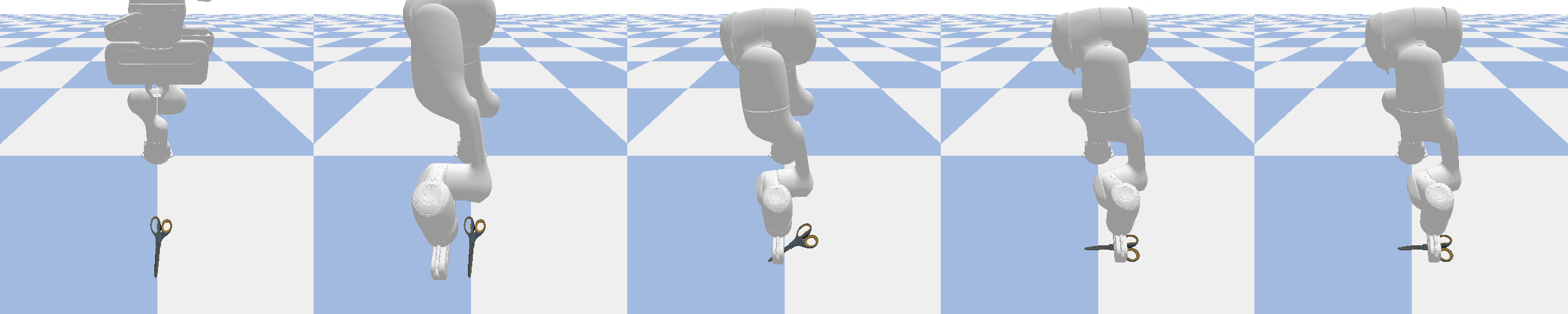} 
    \vspace{-20pt}
\caption{\small Reward update iteration 5: the object is pushed further}
\label{fig:iter_5}
\end{figure}
\begin{figure}[]
\centering
    \includegraphics[width=1\linewidth]{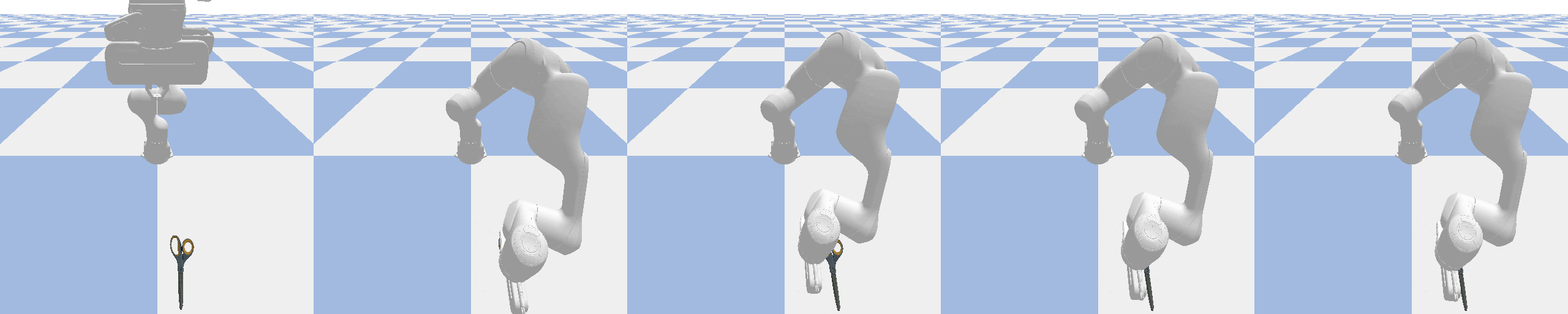} 
    \vspace{-20pt}
\caption{\small Final policy with self-alignment updated reward function: policy learns to establish more stable and consistent pushing pattern}
\label{fig:iter_10}
\end{figure}

\begin{figure}[h]
\centering
    \includegraphics[width=1\linewidth]{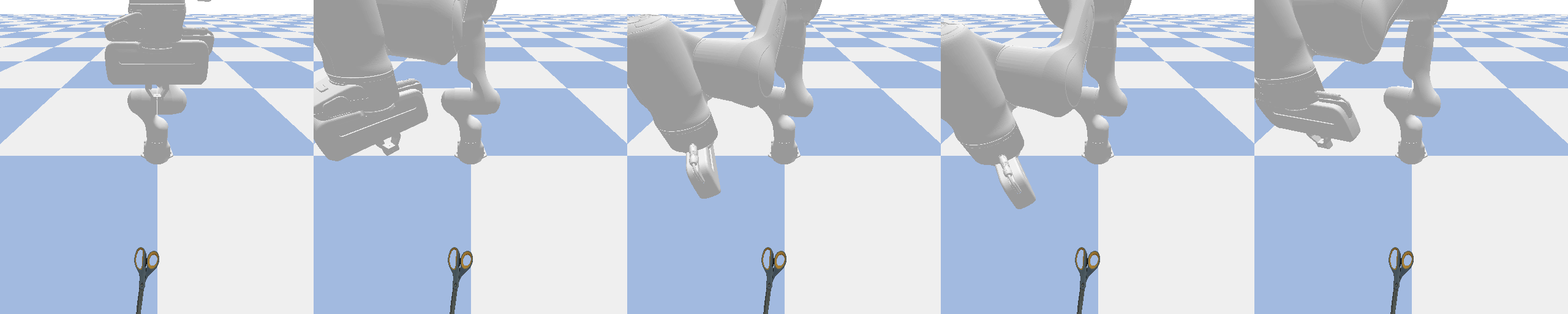} 
    \vspace{-20pt}
\caption{\small Final policy learnt with sparse reward function. The policy only produce random hovering action in the air.}
\label{fig:sparse_final_policy}
\end{figure}

\begin{figure}[h]
\centering
    \includegraphics[width=1\linewidth]{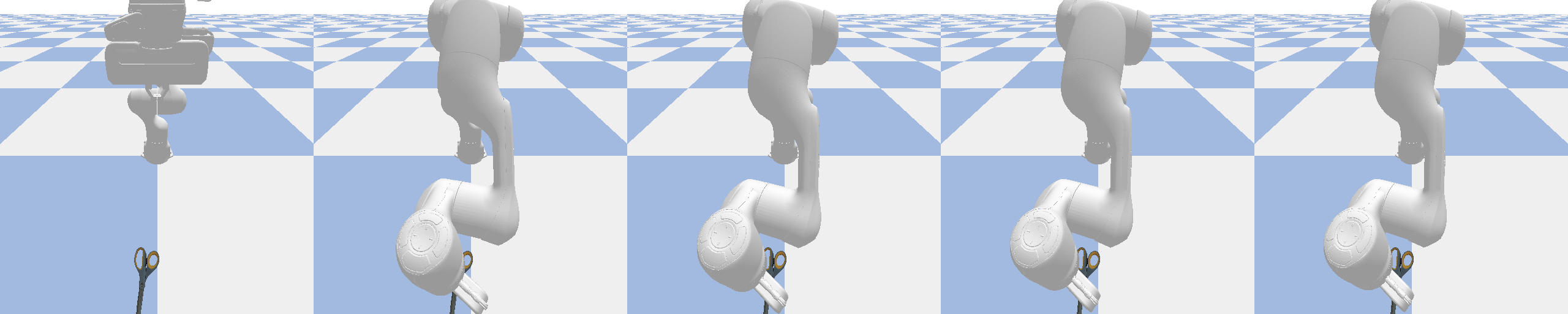} 
    \vspace{-20pt}
\caption{\small Final policy learnt with fixed LLM proposed reward function. The policy learns to reach to the object but fails to make contact.}
\label{fig:fixed_final_policy}
\end{figure}

The same visualization for the final policy over 100,000 steps update using sparse reward and fixed LLM reward function are showed here in figure \ref{fig:sparse_final_policy} and figure \ref{fig:fixed_final_policy}: \\

\end{document}